\documentclass[runningheads]{llncs}
\usepackage[T1]{fontenc}
\usepackage{graphicx,verbatim}
\usepackage{amssymb, bm, amsmath}
\usepackage{tikz}
 \usepackage{booktabs}
\usepackage{multirow} 
\usetikzlibrary{positioning}
\usepackage{subcaption}
\usepackage{pgfplots}
\usepgfplotslibrary{groupplots}
\pgfplotsset{compat=1.18}
\usepackage{soul}

\newif\ifdraft
\drafttrue

\definecolor{orange}{rgb}{1,0.5,0}
\definecolor{gr}{rgb}{0,0.65,0}
\definecolor{mygray}{gray}{0.95}

\ifdraft
 \newcommand{\RS}[1]{{\color{red}{\bf RS: #1}}}
 
 \newcommand{\PMN}[1]{{\color{orange}{\bf PMN: #1}}}
 
 \newcommand{\MCH}[1]{{\color{purple}{\bf MCH: #1}}}
 
\else
 \renewcommand{\sout}[1]{}
 \newcommand{\RS}[1]{{\color{red}{}}}
 
 \newcommand{\PMN}[1]{{\color{red}{}}}
 
 \newcommand{\MCH}[1]{{\color{red}{}}}
 
\fi

\newcommand{\eg}{\textit{e.g.}}
\newcommand{\real}{\mathbb{R}}
\newcommand{\x}{\mathbf{x}}
\newcommand{\z}{\mathbf{z}}

\newcommand{\bv}{\mathbf{v}}

\renewcommand{\H}{\mathbf{H}}
\newcommand{\W}{\mathbf{W}}
\newcommand{\C}{\mathcal{C}}
\newcommand{\Z}{\mathbf{Z}}

\newcommand{\oursubsection}[1]{\textbf{#1.}}

\newcommand*{\XX}{%
  \textsf{X\kern-1ex X}%
}

\begin{document}
%
\title{Beyond Point Estimates for Glaucoma Visual Field Forecasting with Diffusion Models}
\titlerunning{Beyond Point Estimates for Glacoma Visual Fields}
%

\author{Marta Colmenar Herrera\inst{1} \and
Pablo Márquez Neila \inst{1} \and
Şerife Seda Kucur Ergünay\inst{2} \and 
Martin S. Zinkernagel\inst{3}\and
Raphael Sznitman\inst{1} 
}
\authorrunning{M. Colmenar Herrera et al.}
%
\institute{ARTORG Center for Biomedical Engineering Research, UniBe, Switzerland \and
PeriVision SA, Epalinges, Switzerland\and
Department of Ophthalmology, Inselspital, Bern, Switzerland}

  
\maketitle              
\begin{abstract}
Forecasting visual fields (VFs) is critical for personalized monitoring and treatment planning in glaucoma. This is inherently uncertain due to heterogeneous disease progression and measurement variability, yet most existing methods produce single deterministic predictions that fail to represent this uncertainty. We formulate VF forecasting as a probabilistic prediction problem and the use of conditioned denoising diffusion models to generate distributions of plausible future VFs from longitudinal observations with irregular follow-up intervals. Experiments on two independent VF cohorts show that diffusion-based predictions produce well-calibrated distributions for clinically relevant VF measures. When reduced to a standard point-estimate, the proposed approach achieves state-of-the-art accuracy compared to clinical baselines and prior learning-based methods. 
Our results highlight the advantages of distributional modeling for VF forecasting and support a shift from point-estimate prediction toward uncertainty-aware, clinically interpretable risk assessment in glaucoma.





\end{abstract}

\section{Introduction}
Glaucoma is a chronic, progressive optic neuropathy and a leading cause of irreversible blindness worldwide~\cite{GluTham}.
Because progression is typically slow, heterogeneous, and patient-specific, forecasting future functional outcomes remains a long-standing and clinically important challenge~\cite{PraChauhan,PreMedeiros}. Standard Automated Perimetry (SAP) is the primary method for longitudinal functional monitoring. SAP produces visual fields (VFs), which are spatial sensitivity measurements acquired repeatedly to assess disease severity and progression. Clinically, accurate forecasting of future VFs from past measurements can support risk stratification, inform follow-up scheduling, and guide treatment planning~\cite{DetMoraes}.

\begin{figure}[b!]
    \centering
    \begin{subfigure}[t]{0.9\linewidth}
        \centering
        \includegraphics[width=\linewidth]{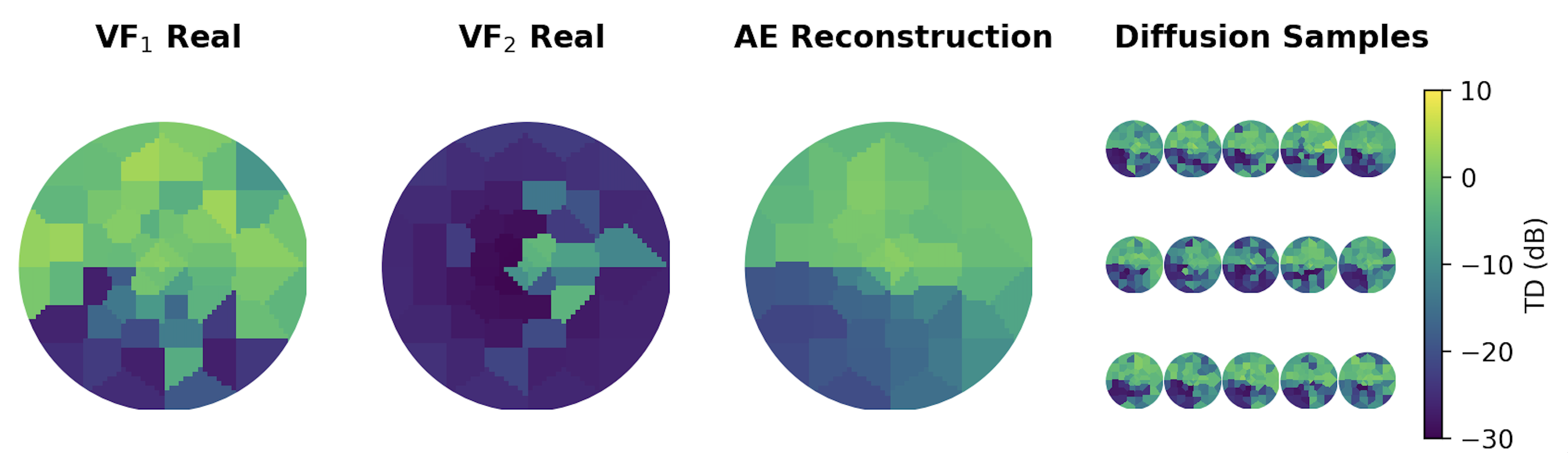}
    \end{subfigure}    
    \begin{subfigure}[t]{0.9\linewidth}
        \centering
        \includegraphics[width=\linewidth]{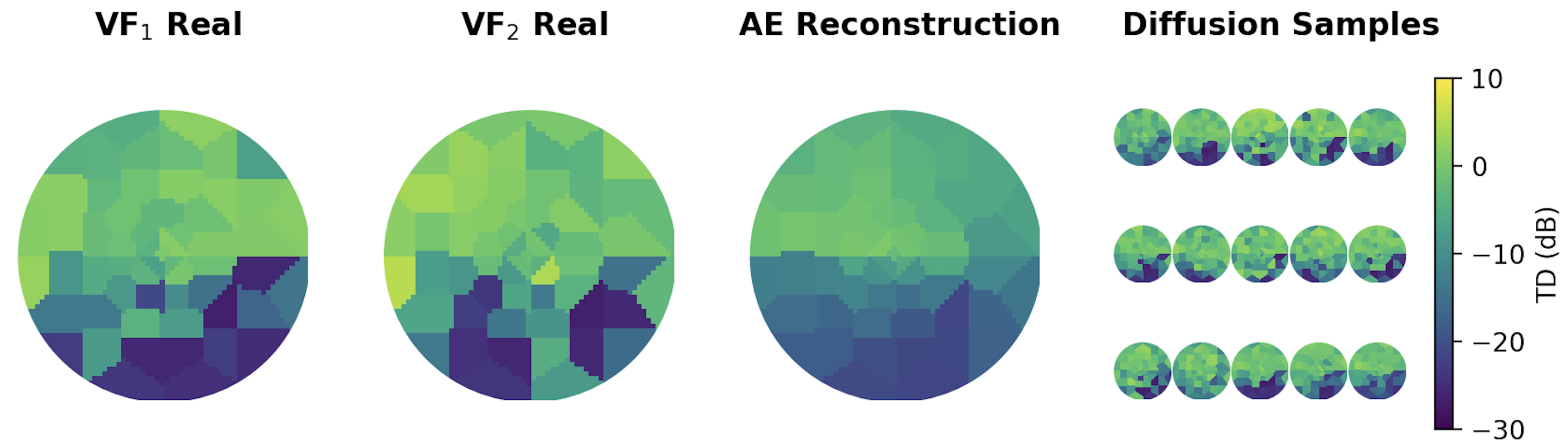}
    \end{subfigure}
    \caption{
    VF forecasting. Two patients with similar VF histories (VF$_1$) exhibit divergent future trajectories~(VF$_2$) (top and bottom row). Deterministic models (\eg.~AE) predict a single smoothed VF corresponding to the conditional mean, which obscures alternative plausible progressions. In contrast, diffusion models generate multiple realistic future VFs, capturing heterogeneous disease evolution.
    }
    \label{fig:similar_trends}
\end{figure}

Despite its clinical importance, VF forecasting is inherently difficult due to the non-deterministic and multi-modal nature of glaucoma progression. Disease trajectories are influenced by latent, partially unobserved factors such as
intraocular pressure, treatment changes, physiological differences, or measurement variability.
As a result, patients with similar historical VFs may experience substantially different future outcomes~\cite{VariabilityTan} (Fig.~\ref{fig:similar_trends}). Consequently, multiple clinically plausible future VFs may be consistent with the same observed history, and a single predicted VF cannot adequately represent this uncertainty.

Traditional VF forecasting approaches rely on simple but interpretable linear trend models, including global index analysis and point-wise linear regression (PLR)~\cite{VFIBengtsson,PLRGardiner}. While efficient and clinically interpretable, these models impose restrictive assumptions and often fail to capture nonlinear progression patterns observed in real-world data~\cite{linearity}. More recently, deep learning approaches based on variational autoencoder (VAE) architectures or recurrent networks (RNN) have improved predictive accuracy by learning complex non-linear spatial dependencies and compact representations~\cite{VAEBerchuck,RNNWen}. However, these methods remain inherently deterministic: for a given set of observed VFs, they produce a single fixed prediction of the future. Such point-estimate predictors implicitly target the conditional mean of the predictive distribution, which may be clinically misleading and can underestimate progression severity in patients with heterogeneous and aggressive disease trajectories. In particular, point-estimate predictors cannot represent the multi-modal structure of the posterior over future VFs.

More recently,~\cite{WenwenDiff} acknowledges uncertainty in VF forecasting by employing generative models with post hoc calibration techniques. However, this approach does not explicitly model disease progression as a stochastic process over time. Instead, it reduces the predictive distribution to a single representative future VF and then quantifies uncertainty around that point prediction using calibrated intervals, thereby collapsing the inherent multi-modality of glaucoma progression and precludes modeling divergent but clinically plausible future trajectories. As a result, VF forecasting remains framed as a one-step prediction problem with uncertainty added post hoc, rather than as longitudinal distributional modeling. 

To address these limitations, VF forecasting methods should move beyond single deterministic or post hoc-calibrated predictions and instead model distributions over future outcomes. We formulate VF forecasting as a probabilistic prediction problem and investigate the use of conditioned diffusion models to directly model the conditional distribution of future VFs given an arbitrary-length history of prior observations and their acquisition times. By explicitly modeling longitudinal uncertainty, our approach captures heterogeneous disease trajectories and generates multiple plausible futures consistent with the same observed history. In practice, our approach can better reflect uncertainty in disease progression, inform personalized follow-up strategies, and identify patients whose future functional loss remains ambiguous despite similar previous measurements. Our experimental results show that the distributions generated by our model are consistent with observed longitudinal VF evolutions.

\section{Method}

We formulate VF forecasting as sampling from a learned conditional distribution over future VFs,
\begin{equation}
p\left(\bv_{N+1} \mid \{(\bv_{i}, t_i)\}_{i=1}^N, t_{N+1}\right),    
\end{equation}
conditioned on $N$~observed VFs,~$\bv_1, \ldots, \bv_N$, acquired at times $0 = t_1 < \dots < t_N$. Time~$t_i$ denotes the elapsed time from $\bv_1$ to $\bv_i$, and $t_{N+1}$ specifies the prediction horizon for the future VF~$\bv_{N+1}$. Each VF~$\bv_i\in\real^L$ is represented as a vector of total deviations~(TD) values at $L$~test locations, expressed 
in dB relative to age-corrected sensitivities. Negative values indicate localized functional loss.

\oursubsection{Conditional diffusion model}
We model the conditional distribution using a standard conditioned denoising diffusion probabilistic model (cDDPM)~\cite{DDPM}, where the target future VF~$\bv_{N+1}$ corresponds to the final denoised sample~$\x_0$. The reverse diffusion process is parameterized by a noise-prediction network~$\epsilon_\theta$ following the standard cDDPM formulation. 


\oursubsection{Noise-prediction network}
The noise-prediction network~$\epsilon_\theta$ is based on a transformer architecture with self-attention and cross-attention conditioning blocks. At each denoising step~$\tau \in \{1,\ldots,T\}$, the network takes as input the noisy diffusion state~$\x_\tau$ 
with the observed VF history~$\C=\left\{\{\bv_i,t_i\}_{i=1}^N, t_{N+1}\right\}$ and the diffusion step~$\tau$. Both the noisy state and the conditioning VFs are embedded into a sequence of per-location tokens encoding TD values, the spatial location~$\ell$, and the acquisition time~$t_i$. For an observed VF~$\bv_i$, each location is embedded as
$
    \z_{i\ell}
    =
    \W_{\text{val}}\,\phi(\bv_{i\ell})
    +
    \W_{\text{time}}\,\phi(t_i)
    +
    \W_{\text{space}}\,\phi(\ell),
$
where $\phi$ denotes a sinusoidal encoding function, and $\W_{\text{val}}$, $\W_{\text{time}}$, and $\W_{\text{space}}$ are learned linear projections. All conditioning tokens are concatenated in a matrix~$\Z\in\real^{L\cdot N\times D}$. The noisy diffusion state is embedded analogously into state tokens~$\H_\tau\in\real^{L\times D}$. The state tokens are processed by $K$~stacked transformer blocks, each comprising self-attention among state tokens~$\H_\tau$ followed by cross-attention to the conditioning tokens~$\Z$, with residual connections and normalization. After $K$ blocks, a shared linear layer maps each final state token to a scalar noise prediction.

\oursubsection{Training}
We train the model using the standard denoising diffusion objective, minimizing the MSE between injected noise and predicted noise. Training samples are obtained from consecutive VF sequences, with the first~$N$ VFs and acquisition times forming the conditioning information~$\mathcal{C}$ and the subsequent VF as the target. Optimization is performed via stochastic gradient descent with random diffusion steps and noise realizations.


\oursubsection{Inference}
At inference time, future VFs are generated by sampling from the learned reverse diffusion process conditioned on the observed VF history~$\mathcal{C}$. Sampling is initialized from $\x_T \sim \mathcal{N}(\mathbf{0},\mathbf{I})$ and proceeds iteratively from $\tau=T$ to~$\tau=1$, drawing $\x_{\tau-1} \sim p_\theta(\x_{\tau-1} \mid \x_\tau, \mathcal{C})$ using the noise-prediction network~$\epsilon_\theta(\x_\tau,\mathcal{C},\tau)$. The final sample $\x_0$ corresponds to a generated future VF~$\hat{\bv}_{N+1}$.

Repeating this procedure yields $S$~independent samples for the same conditioning, forming an empirical approximation of the conditional predictive distribution. This representation enables uncertainty-aware inference, such as predictive intervals or probabilities of clinically relevant outcomes, rather than a single point estimate. The number of samples~$S$ controls the trade-off between approximation accuracy and computational cost.

\begin{table*}[b!]
    \centering
    {\fontsize{8}{9}\selectfont
    \begin{tabular}{lcccccc}\toprule
        \multirow{2}{*}{Dataset} & No. & No.&No. VFs/Subject&Follow-up& Age (Mean$\pm$SD & MD (Mean$\pm$SD  \\
        & Subjects & VFs&(Mean$\pm$SD)&duration& [Min, Max])& [Min, Max] dB)\\
        \midrule
        \multirow{2}{*}{ICVF} & \multirow{2}{*}{$8,463$} & \multirow{2}{*}{$38,832$} &   $2.36\pm3.01$ &$1.48\pm 3.36$&$57.49\pm17.83$ & $-4.45\pm5.61$  \\
         & & &   $[1.00,35.00]$&$[0.00, 24.22]$&$[7.22, 95.57]$ & $[-28.90,9.20]$  \\
         
        \multirow{2}{*}{UWHVF} & \multirow{2}{*}{$3,867$} & \multirow{2}{*}{$28,553$} &   $3.85\pm2.53$&$3.88\pm 3.93$&$64.50\pm14.46$ & $-6.94\pm6.18$  \\
        & & &   $[1.00,19.00]$&$[0.00, 20.81]$&$[10.06, 90.00]$ & $[-32.95, 0.00]$  \\ \bottomrule
    \end{tabular}}
    \caption{Dataset characteristics for the private clinical cohort (ICVF, OCTOPUS) and public cohort (UWHVF, HFA), including age and MD statistics.}
    \label{tab:demographics_datasets}
\end{table*}

\section{Experiments}

We evaluated our approach in two ways: (1) predictive accuracy compared to point-estimate baselines, and (2) fidelity of the predicted distributions with respect to observed longitudinal VF outcomes.

\oursubsection{Datasets}
We evaluated our method using two independent longitudinal VF cohorts: an internal clinical VF dataset~(ICVF) and a publicly available dataset~(UWHVF)~\cite{UWHVF}, acquired with the OCTOPUS 900 and the Humphrey Field Analyzer II (HFA), respectively (Table~\ref{tab:demographics_datasets}). Standard quality control excluded VFs with false-positive or false-negative rates above 30\%, as well as fields exhibiting non-glaucomatous defect patterns as determined by established clinical criteria. Each dataset consists of patient longitudinal VF sequences, along with their associated acquisition times. Subject-level splits were used to assign subjects to training and test sets, with approximately~80\% of VFs used for training and~20\% for testing. Instances were constructed by sliding a temporal window over each subject’s sequence of consecutive VFs to form tuples~$(\bv_1, \ldots, \bv_N, \bv_{N+1})$. To account for variable history lengths, we extracted instances using multiple values of~$N\in\{1,2,3\}$.

\begin{table}[b!]
\centering
{\fontsize{8}{9}\selectfont
\setlength{\tabcolsep}{1pt}
\begin{tabular}{l|cccc|cccc}
\toprule
& \multicolumn{4}{c|}{$N=1$} & \multicolumn{4}{c}{$N=2$} \\
\cmidrule(lr){2-5} \cmidrule(lr){6-9} & \multirow{2}{*}{Ours}& CasNet-5~& VAE& No-change& \multirow{2}{*}{Ours}& CasNet-5& VAE& PLR \\
Subset& & \cite{RNNWen}& ~\cite{VAEBerchuck}& predictor& & \cite{RNNWen}& ~\cite{VAEBerchuck}&\\
\midrule

\multirow{2}{*}{\textbf{ICVF (758)}}& \textbf{3.08} & 3.12 & 3.37 & 3.33 
& \textbf{2.89} & 2.97 & 3.59 & 10.83 \\
& \textbf{$\pm$1.75} & $\pm$1.79 & $\pm$1.96 & $\pm$1.74 
& \textbf{$\pm$1.67} & $\pm$1.65 & $\pm$2.03 & $\pm$25.17 \\    

\multirow{2}{*}{Earliest (57)}
& 1.89 & \textbf{1.83} & 2.41 & 2.27 
& \textbf{1.79} & 2.31 & 2.33 & 7.28 \\
& $\pm$0.83 & \textbf{$\pm$0.86} & $\pm$1.26 & $\pm$1.05 
& \textbf{$\pm$0.76} & $\pm$1.79 & $\pm$1.01 & $\pm$11.91 \\

\multirow{2}{*}{Early (429)}
& \textbf{2.30} & 2.33 & 2.35 & 2.64 
& 2.25 & \textbf{2.16} & 2.65 & 9.44 \\
& \textbf{$\pm$1.16} & $\pm$1.19 & $\pm$1.15 & $\pm$1.30 
& $\pm$1.17 & \textbf{$\pm$0.82} & $\pm$1.42 & $\pm$26.15 \\

\multirow{2}{*}{Moderate (161)}
& \textbf{4.15} & 4.27 & 4.40 & 4.38 
& \textbf{3.88} & 3.92 & 4.77 & 11.12 \\
& \textbf{$\pm$1.64} & $\pm$1.54 & $\pm$1.30 & $\pm$1.63 
& \textbf{$\pm$1.51} & $\pm$1.15 & $\pm$1.74 & $\pm$16.44 \\

\multirow{2}{*}{Advanced (82)}
& \textbf{5.20} & 5.31 & 5.97 & 5.27 
& \textbf{4.78} & 4.79 & 6.03 & 18.36 \\
& \textbf{$\pm$1.76} & $\pm$1.78 & $\pm$1.39 & $\pm$1.90 
& \textbf{$\pm$1.70} & $\pm$1.47 & $\pm$1.57 & $\pm$36.81 \\

\multirow{2}{*}{Severe (29)}
& 5.02 & 5.15 & 7.16 & \textbf{4.54} 
& \textbf{4.62} & 5.56 & 6.50 & 15.60 \\
& $\pm$1.00 & $\pm$1.05 & $\pm$2.43 & \textbf{$\pm$1.28} 
& \textbf{$\pm$1.08} & $\pm$2.83 & 1.73 & $\pm$24.57 \\
\midrule

\multirow{2}{*}{\textbf{UWHVF (702)}}& \textbf{2.62} & 2.69 & 3.10 & 2.92 
& \textbf{2.51} & 2.58 & 3.20 & 6.46 \\
& \textbf{$\pm$1.71} & $\pm$1.75 & $\pm$2.04 & $\pm$1.84 
& \textbf{$\pm$1.63} & $\pm$1.62 & $\pm$2.09 & $\pm$7.46 \\

\multirow{2}{*}{Early (398)}
& \textbf{1.86} & 1.91 & 2.07 & 2.16 
& \textbf{1.80} & 1.82 & 2.24 & 4.63 \\
& \textbf{$\pm$1.08} & $\pm$1.05 & $\pm$1.06 & $\pm$1.17 
& \textbf{$\pm$1.05} & $\pm$1.59 & $\pm$1.17 & $\pm$3.94 \\

\multirow{2}{*}{Moderate (231)}
& \textbf{3.24} & 3.36 & 3.78 & 3.62 
& 3.14 & \textbf{3.00} & 4.06 & 8.56 \\
& \textbf{$\pm$1.74} & $\pm$1.89 & $\pm$1.74 & $\pm$1.86 
& $\pm$1.76 & \textbf{$\pm$1.29} & $\pm$2.28 & $\pm$10.29 \\

\multirow{2}{*}{Advanced (52)}
& \textbf{5.01} & 5.09 & 6.43 & 5.12 
& \textbf{4.65} & 5.19 & 6.05 & 10.87 \\
& \textbf{$\pm$1.47} & $\pm$1.44 & $\pm$1.75 & $\pm$1.67 
& \textbf{$\pm$1.49} & $\pm$1.74 & $\pm$2.15 & $\pm$9.41 \\

\multirow{2}{*}{Severe (21)}
& \textbf{4.16} & 4.27 & 6.79 & 4.18 
& \textbf{3.59} & 4.85 & 4.82 & 7.22 \\
& \textbf{$\pm$2.33} & $\pm$2.55 & $\pm$3.39 & $\pm$3.37 
& \textbf{$\pm$1.52} & $\pm$3.35 & $\pm$1.68 & $\pm$4.84 \\
\bottomrule
\end{tabular}}
\caption{Mean absolute error (MAE; mean$\pm$SD) for point-estimate future VF prediction using $N=1$ and $N=2$ observed VFs. Results are reported for the full test set and stratified by glaucoma severity stage (sample sizes in parentheses).}
\label{tab:baseline_comparison2}
\end{table}

\begin{figure}[b!]
    \centering
    \includegraphics[width=0.92\linewidth]{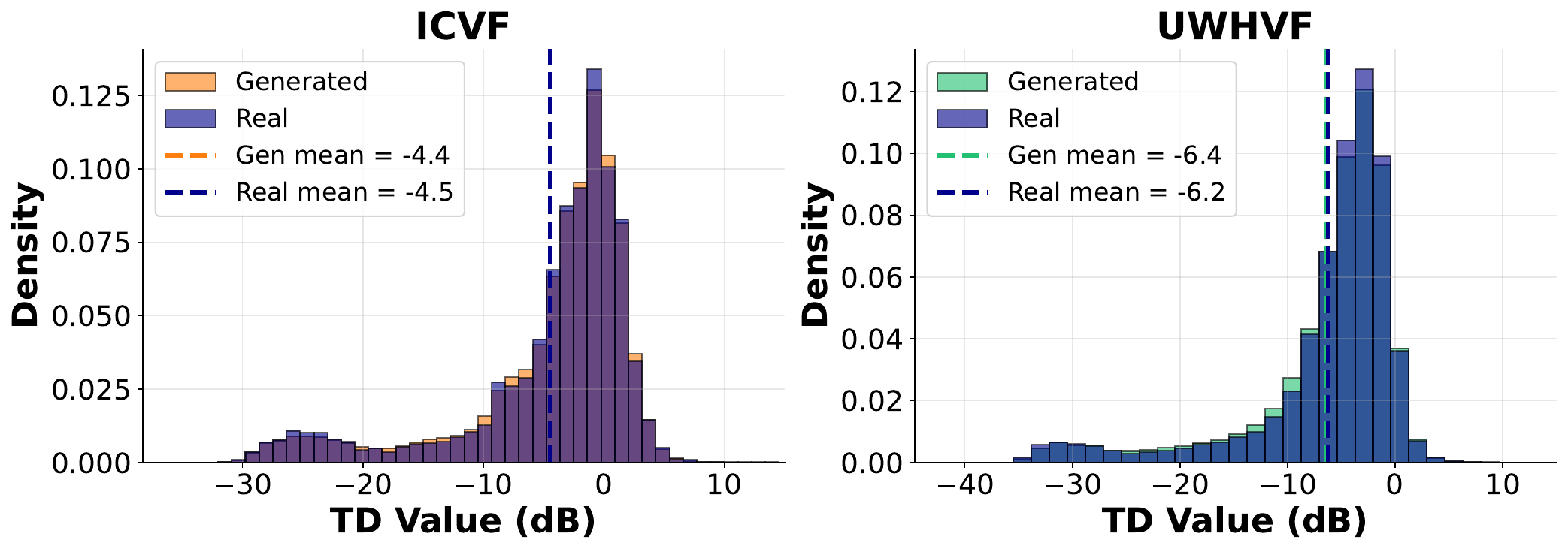}
    \caption{Marginal TD distributions of predicted and observed future VFs.}
    \label{fig:TD_distribution}
\end{figure}

To enable stratified analysis by disease severity, we adopted a device-specific staging classification~\cite{linearity}, which categorizes VFs into five stages (earliest, early, moderate, advanced, and severe) based on the VF mean defect~(MD). MD is a global index computing as the average of TD values across all test locations~\cite{Digest}.



\oursubsection{Implementation details}
Models were trained using AdamW with learning rate~$\eta=10^{-4}$ and batch size~512 for~15{,}000~epochs. The embedding dimension was set to~$D=64$, with~$K=2$ transformer blocks. The number of VF locations was~$L=59$ for ICVF and~$L=52$ for UWHVF. The diffusion process used~$T=52$ denoising steps. Model checkpoints were evaluated every~50~epochs on a held-out validation set comprising~15\% of the training data. Checkpoints were compared using the Fréchet distance between multivariate Gaussian fits to real future VFs and to generated samples, used solely for model selection. All experiments were conducted on NVIDIA RTX~4090 GPUs. During inference, $S=2000$ independent future VF samples were generated per test instance.



\oursubsection{Point-estimate baselines}
We compared our method against baselines that produce single deterministic VF predictions to assess point-wise accuracy under standard evaluation metrics and to enable comparison with prior work:
\begin{description}
    \item[No-change.] A persistence baseline predicting the last observed VF, $\hat{\bv}_{N+1}=\bv_N$, serving as a simple reference.
    
    \item[PLR.] Point-wise linear regression, reflecting standard clinical practice, fitted independently at each VF location and extrapolated to~$t_{N+1}$.
    
    \item[VAE.] A variational autoencoder-based VF forecasting model~\cite{VAEBerchuck}.
    
    \item[CascadeNet-5.] An RNN model for longitudinal VF prediction~\cite{RNNWen}.
    
\end{description}
All learning-based baselines were trained using mean squared error (MSE) to predict future VFs from observed VF sequences and acquisition times.

Point-estimate forecasting performance was evaluated using mean absolute error (MAE) between predicted and observed future VFs. Since our method produces a predictive distribution, point estimates were obtained using the predictive mean, computed by averaging generated samples, which is consistent with MSE-based approaches.


\oursubsection{Point-estimate prediction performance}
Even when reduced to a point estimate using the predictive mean, our method achieves the lowest MAE across all settings, outperforming clinical baselines and prior learning-based approaches (Table~\ref{tab:baseline_comparison2}). Improvements are most pronounced in moderate and advanced stages, where VF variability is higher and forecasting is more challenging~\cite{VarGardiner}. In earliest and early stages, VFs are relatively stable and performance differences between methods are smaller. Overall, these results show that explicitly modeling uncertainty does not compromise point-estimate accuracy and can yield more accurate predictions than methods optimized solely for deterministic forecasting.
\begin{figure}[b!]
    \centering
    \includegraphics[width=0.90\linewidth]{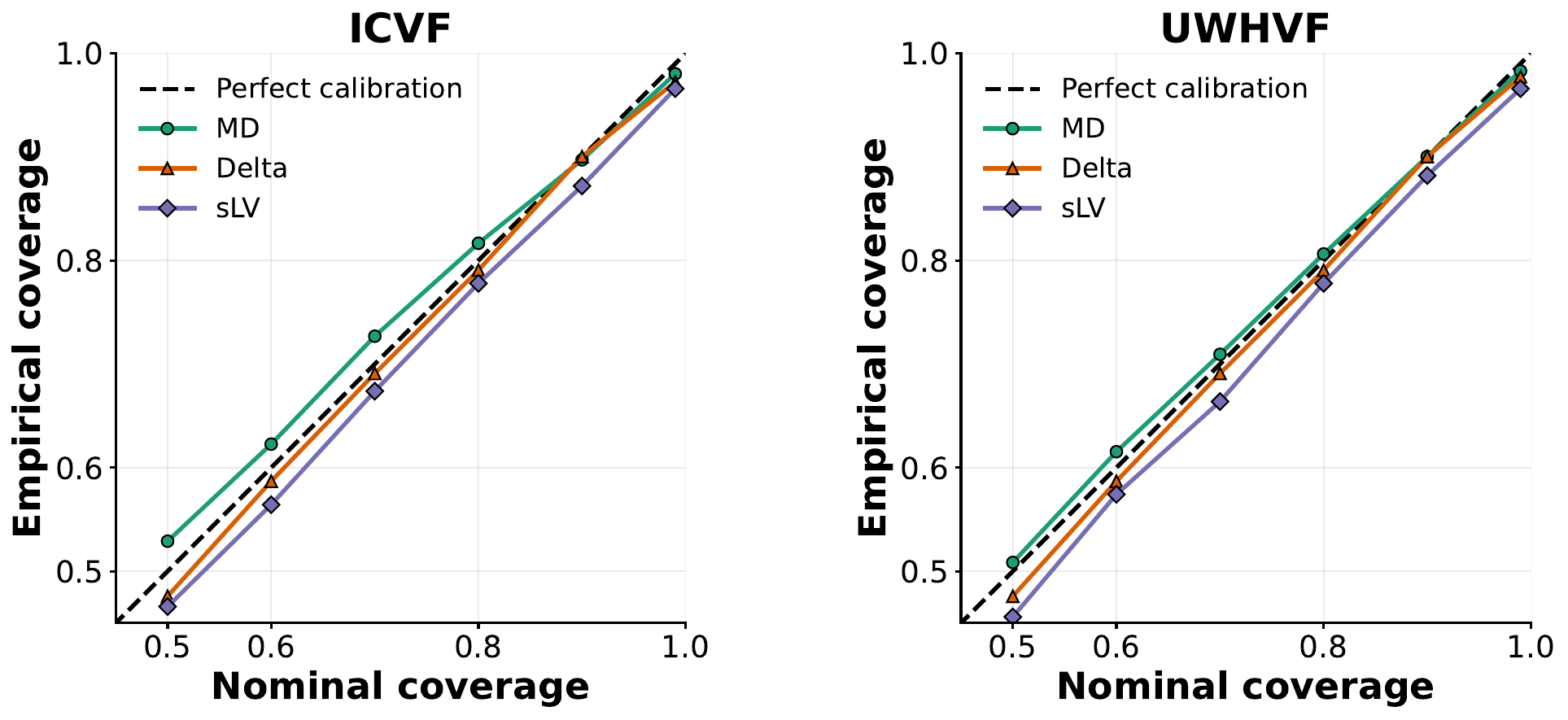}
    \caption{Calibration curves comparing nominal and empirical coverage for two datasets. The dashed line ($y=x$) indicates perfect calibration. }
    \label{fig:calibration}
\end{figure}

\begin{figure}[b!]
    \centering
    \includegraphics[width=0.90\linewidth]{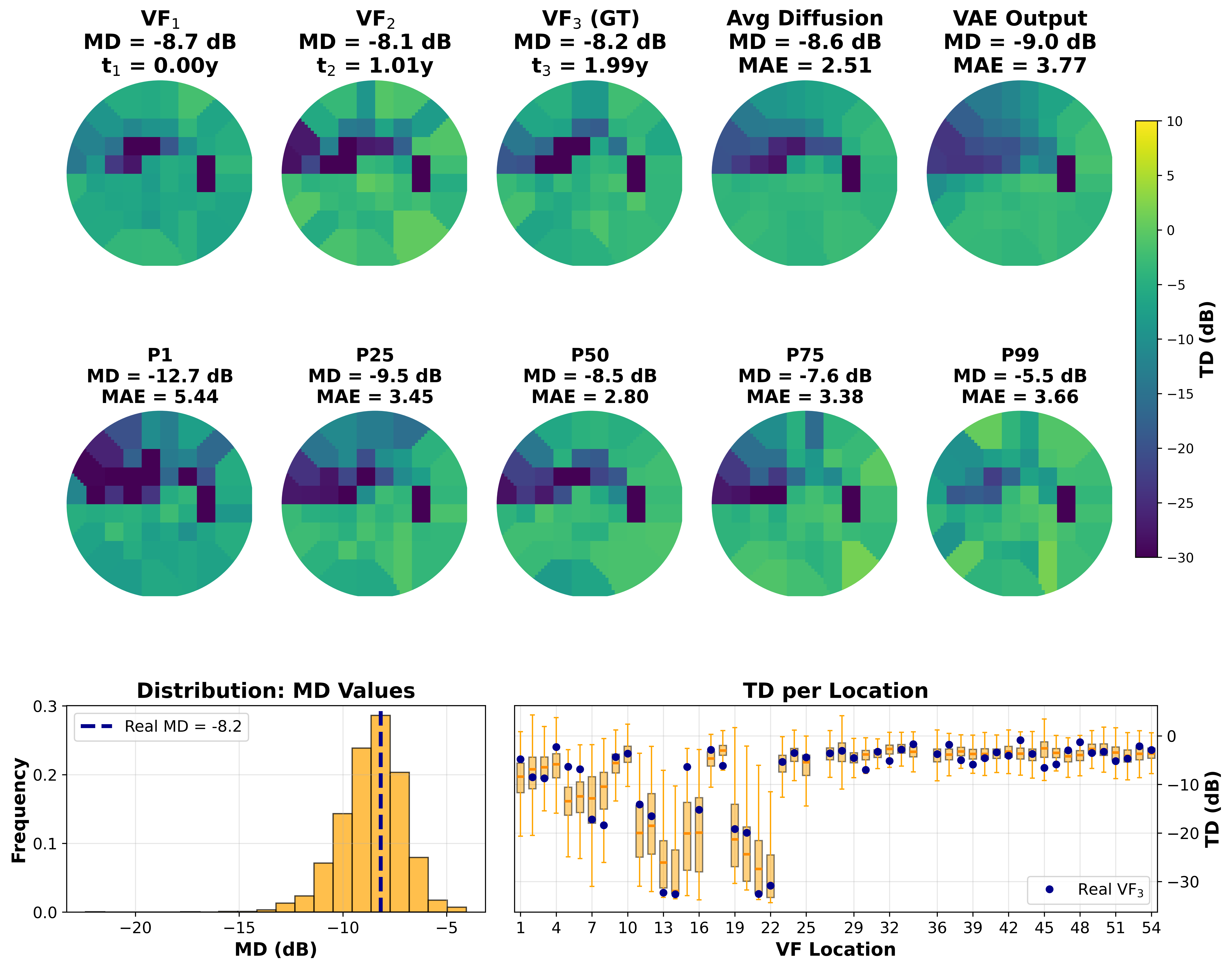}
    \caption{Qualitative uncertainty visualization for a UWHVF test sample. Top rows show two observed VFs (VF$_1$ and VF$_2$), the ground-truth future VF (VF$_3$), the predictive mean of the diffusion model, the VAE baseline prediction, and five diffusion samples selected at the 1st, 25th, 50th, 75th, and 99th percentiles of predicted MD. Bottom panels show the empirical distribution of predicted MD values and per-location TD uncertainty
    , with true values indicated in blue.
    }
    \label{fig:qualitative_example}
\end{figure}

\oursubsection{Predictive distribution analysis}
As no prior methods provide predictive distributions for VF forecasting, we evaluate distributional performance through calibration, marginal consistency, and qualitative case studies.

\textbf{Calibration.} Calibration curves assess the agreement between predicted probabilistic forecasts and observed outcomes and are essential for determining whether predicted uncertainties can be interpreted as reliable probabilities. We evaluate calibration on three clinically relevant VF summary measures: mean defect~(MD), square root of loss variance~(sLV)~\cite{Digest}, and local defect magnitude~($\Delta$)~\cite{delta,delta2}, defined as the maximum absolute difference between neighboring test locations. For each metric and nominal coverage~$\alpha$, we construct centered predictive intervals defined by the quantiles~$[Q_{(1-\alpha)/2}, Q_{(1+\alpha)/2}]$ of the predicted distribution for each test case. Empirical coverage is then computed as the proportion of test cases for which the observed future measure falls within the predictive interval. Well-calibrated predictive distributions yield empirical coverage values close to the nominal level across different~$\alpha$. 

Calibration curves show close agreement between nominal and empirical coverage across all considered measures and datasets, indicating that the predicted distributions accurately quantify uncertainty (Fig.~\ref{fig:calibration}). This constitutes a key finding of this study, as calibration is essential for ensuring that predicted probabilities can be interpreted as meaningful measures of clinical risk. Because the predictive distributions are well calibrated for clinically relevant VF measures, probabilistic queries --—such as estimating the probability that a patient’s MD will cross a clinical threshold—-- can be answered directly from the predicted distributions. The proposed diffusion model enables such probabilistic queries without requiring additional post hoc calibration.


\textbf{Marginal consistency.} We also evaluate marginal consistency by comparing the distributions of predicted and observed future outcomes aggregated over the test set. The predicted marginal distribution is obtained by merging the conditional predictions of the diffusion model across test instances. This comparison assesses whether the model reproduces population-level statistics of VF progression. We observe that samples from the predictive distributions closely match the observed distribution of TD 
in the test set, capturing heavy-tailed defect behavior in TD values~(Fig.~\ref{fig:TD_distribution}) 

\textbf{Qualitative visualization.}
Distributional forecasting enables explicit representation of uncertainty in future VF outcomes. In a representative example (Fig.~\ref{fig:qualitative_example}), the diffusion model generates multiple plausible future fields consistent with the observed history, capturing both global severity variation and localized defect patterns. The observed future VF lies within the support of the predicted distribution, whereas a deterministic baseline produces a single smoothed outcome that cannot represent alternative trajectories.

\section{Conclusion}
We presented a diffusion-based framework for probabilistic prediction of future VFs from longitudinal perimetry data. We show in this work that the proposed model generates well-calibrated predictive distributions for clinically relevant VF measures without requiring post hoc calibration, while also achieving state-of-the-art performance when reduced to a point estimate via the predictive mean. Our findings highlight the limitations of purely point-estimate forecasting methods, which cannot adequately represent uncertainty or support risk-aware interpretation. In contrast, probabilistic forecasts enable clinically meaningful queries (\eg,~``what is the probability of MD dropping by 5 points in 6 months?''). 
Our results suggest that probabilistic, distributional modeling provides a more informative and clinically aligned paradigm for future research in VF forecasting.


\begin{thebibliography}{8}
\bibitem{GluTham}
Tham, Y. C., Li, X., Wong, T. Y., Quigley, H. A., Aung, T., Cheng, C. Y. Global prevalence of glaucoma and projections of glaucoma burden through 2040: a systematic review and meta-analysis. Ophthalmology, \textbf{121}(11), 2081--2090 (2014)

\bibitem{PraChauhan}
Chauhan, B. C., Garway-Heath, D. F., Goñi, F. J., Rossetti, L., Bengtsson, B., Viswanathan, A. C., Heijl, A. Practical recommendations for measuring rates of visual field change in glaucoma. The British journal of ophthalmology, \textbf{92}(4), 569--573 (2008)

\bibitem{PreMedeiros}
Medeiros, F. A., Alencar, L. M., Zangwill, L. M., Bowd, C., Sample, P. A., Weinreb, R. N. Prediction of functional loss in glaucoma from progressive optic disc damage. Archives of ophthalmology, \textbf{127}(10), 1250--1256 (2009)

\bibitem{DetMoraes}
De Moraes, C. G., Liebmann, J. M., Levin, L. A. Detection and measurement of clinically meaningful visual field progression in clinical trials for glaucoma. Progress in retinal and eye research, \textbf{56}, 107--147 (2017)

\bibitem{VariabilityTan}
Tan, J., Agar, A., Kalloniatis, M., Phu, J. Quantification and Predictors of Visual Field Variability in Healthy, Glaucoma Suspect, and Glaucomatous Eyes Using SITA-Faster. Ophthalmology, \textbf{131}(6), 658--666 (2024)

\bibitem{VFIBengtsson}
Bengtsson, B., Heijl, A. A visual field index for calculation of glaucoma rate of progression. American journal of ophthalmology, \textbf{145}(2), 343--353 (2008)

\bibitem{PLRGardiner}
Gardiner, S. K., Crabb, D. P. Examination of different pointwise linear regression methods for determining visual field progression. Investigative ophthalmology \& visual science, \textbf{43}(5), 1400--1407 (2002)

\bibitem{linearity}
Colmenar Herrera, M., Kucur, Ș., Márquez Neila, P., Sznitman, R. Unterlauft, J. Visual field-based machine learning model for predicting disease stage in glaucoma. Current Directions in Biomedical Engineering, \textbf{11}(1), 366--369 (2025)

\bibitem{VAEBerchuck}
Berchuck, S. I., Mukherjee, S., Medeiros, F. A. Estimating Rates of Progression and Predicting Future Visual Fields in Glaucoma Using a Deep Variational Autoencoder. Scientific reports, \textbf{9}(1), 18113 (2019)   

\bibitem{RNNWen}
Wen, J. C., Lee, C. S., Keane, P. A., Xiao, S., Rokem, A. S., Chen, P.P., Wu, Y., Lee, A. Y. Forecasting future Humphrey Visual Fields using deep learning. PLOS ONE \textbf{14}(4): e0214875 (2019) 

\bibitem{WenwenDiff}
Si, W. et al. (2026). Reliable and Interpretable Visual Field Progression Prediction with Diffusion Models and Conformal Risk Control. In: Gee, J.C., et al. Medical Image Computing and Computer Assisted Intervention – MICCAI 2025. MICCAI 2025. Lecture Notes in Computer Science, vol 15974. Springer, Cham.

\bibitem{DDPM} 
Ho, J., Jain, A., Abbeel, P., Denoising Diffusion Probabilistic Models. arXiv (2020)

\bibitem{UWHVF}
Montesano, G., Chen, A., Lu, R., Lee, C.S., Lee, A.Y. UWHVF: A Real-World, Open Source Dataset of Perimetry Tests From the Humphrey Field Analyzer at the University of Washington. Translational Vision Science \& Technology \textbf{11}(1):2 (2022).

\bibitem{Digest}
Racette, L., Fischer, M., Bebie, H., Holló, G., Johnson, C.A., Matsumoto, C: Automated Perimetry Visual Field Digest. 6th ed. Haag-Streit AG, Switzerland (2016). 

\bibitem{VarGardiner}
Gardiner, S.K., Swanson, W.H., Mansberger, S.L.; Long- and Short-Term Variability of Perimetry in Glaucoma. Translational Vision Science \& Technology \textbf{11}(8):3 (2022).

\bibitem{delta}
Chong, L.X., McKendrick, A.M., Ganeshrao, S.B., Turpin, A. Customized, Automated Stimulus Location Choice for Assessment of Visual Field Defects. Investigative ophthalmology \& visual science \textbf{55}(5):3265--3274 (2014).

\bibitem{delta2}
Kucur, S.S., Häckel, S., Stapelfeldt, J., Odermatt, J., Iliev, M.E., Abegg, M., Sznitman, R., Höhn, R. Comparative Study Between the SORS and Dynamic Strategy Visual Field Testing Methods on Glaucomatous and Healthy Subjects. Translational Vision Science \& Technology \textbf{9}(13):3 (2020).











\end{thebibliography}
\end{document}